\pdfoutput=1
\documentclass[letterpaper]{article} 
\usepackage{aaai2026}  
\usepackage{times}  
\usepackage{helvet}  
\usepackage{courier}  
\usepackage[hyphens]{url}  
\usepackage{graphicx} 
\usepackage{natbib}  
\usepackage{caption} 
\usepackage{algorithm}
\usepackage{algpseudocode}
\usepackage{xspace}
\usepackage{amsmath}
\usepackage{adjustbox}
\usepackage{multirow}
\usepackage{booktabs}
\usepackage{tabularx}
\usepackage{arydshln}   
\usepackage{amssymb}    
\usepackage{animate}
\usepackage{enumitem}
\usepackage{graphicx}
\usepackage{animate}
\usepackage{array}
\graphicspath{{rebuttal_videos/image_pca/}{rebuttal_videos/image_frames/}{rebuttal_videos/video_pca/}{rebuttal_videos/video_frames/}{rebuttal_videos/jumprope/}{rebuttal_videos/jumprope_pca/}{videos/row_1/}{videos/row_2/}{videos/row_3/}
{videos/row_4/}{videos/row_5/}{videos/video_ffs_origin/}{videos/video_ffs_ours/}{videos/video_origin/}{videos/video_ours/}{videos/video_sky_origin/}{videos/video_sky_ours/}}
\newcommand{\framework}{Align4Gen\xspace}
\def\vx{{\boldsymbol{x}}}
\def\x{{\boldsymbol{x}}}

\usepackage{newfloat}
\usepackage{listings}
\DeclareCaptionStyle{ruled}{labelfont=normalfont,labelsep=colon,strut=off} 
\floatstyle{ruled}
\newfloat{listing}{tb}{lst}{}
\floatname{listing}{Listing}
\title{Improving Video Diffusion Transformer Training by Multi-Feature Fusion and Alignment from Self-Supervised Vision Encoders}
\author {
    Dohun Lee\equalcontrib\textsuperscript{\rm 1},
    Hyeonho Jeong\equalcontrib\textsuperscript{\rm 1, \rm 2},
    Jiwook Kim\textsuperscript{\rm 1},
    Duygu Ceylan\textsuperscript{\rm 2},
    Jong Chul Ye\textsuperscript{\rm 1}
}
\affiliations {
    \textsuperscript{\rm 1}KAIST AI\\
    \textsuperscript{\rm 2}Adobe Research \\
    leedh7@kaist.ac.kr, hyeonhoj@adobe.com, tom919@kaist.ac.kr, ceylan@adobe.com, jong.ye@kaist.ac.kr
}

\begin{document}

\maketitle

\begin{abstract}
Video diffusion models have advanced rapidly in the recent years as a result of series of architectural innovations (e.g., diffusion transformers) and use of novel training objectives (e.g., flow matching). In contrast, less attention has been paid to improving the feature representation power of such models. In this work, we show that training video diffusion models can benefit from aligning the intermediate features of the video generator with feature representations of pre-trained vision encoders. We propose a new metric and conduct an in-depth analysis of various vision encoders to evaluate their discriminability and temporal consistency, thereby assessing their suitability for video feature alignment. 
Based on the analysis, we present \framework which provides a novel multi-feature fusion and alignment method integrated into video diffusion model training. We evaluate \framework both for unconditional and class-conditional video generation tasks and show that it results in improved video generation as quantified by various metrics. Full video results are available on our project page: \textit{https://align4gen.github.io/align4gen/} 
\end{abstract}


\section{Introduction}
Video generative models have made rapid strides in recent years.
State-of-the-art video diffusion models \cite{polyak2024moviegencastmedia, videoworldsimulators2024, kong2024hunyuanvideo, yang2024cogvideox, ma2025stepvideot2vtechnicalreportpractice} have advanced in synthesizing high-resolution photorealistic videos that last several minutes. 
To achieve these advancements, recent studies have focused on architectural innovations, e.g., transitioning from U-Net to Diffusion Transformers (DiT) \cite{peebles2023scalable}, and rethinking the training objective, such as flow matching \cite{lipman2022flow}.

In contrast, less effort has been made to analyze the correlation between the feature representation power of large video models and their generation capabilities. In fact, various recent works have demonstrated that large-scale diffusion models learn powerful and discriminative features that can facilitate various analysis tasks \cite{tang2023emergent, luo2023diffusion, zhang2023tale}. 
Such works make two important observations: first, the features of generative models are still inferior~\cite{elbanani2024probing, jeong2024track4gen} compared to the features of self-supervised vision models, such as DINOv2~\cite{oquab2023dinov2}. More interestingly, the features of the generative models and self-supervised vision models tend to exhibit complementary properties~\cite{zhang2023tale}. Inspired by these observations, 
REPA~\cite{yu2024representation} demonstrated that aligning the internal representations of DiT with an external pre-trained visual encoder during training significantly improves both discriminative and generative performance.

Given the success of REPA,
we hypothesize that pre-trained self-supervised vision models can be used as guidance when training video diffusion models to ensure they learn more discriminative features. 
Moreover, we assume that operating at a more discriminative feature space would facilitate higher quality video generation.
Consequently, in this work, we introduce \framework, a novel framework that exploits the rich representation of pre-trained vision encoders to improve video diffusion training. 
We begin by proposing a novel metric that quantifies both discriminability and temporal invariance, and use it to analyze features from a variety of pre-trained image and video encoders.
Surprisingly, image-trained encoders often exhibit superior discriminability and temporal consistency.
We further observe that different pre-trained image features, such as DINOv2~\cite{oquab2023dinov2} and SAM2.1 Hiera~\cite{ravi2024sam2}, focus on capturing different frequency signals and hence \textit{exhibit complementary characteristics}. 
we design a multi-feature fusion and alignment strategy that aligns intermediate features of video diffusion transformers with those from multiple pre-trained image encoders.
We validate \framework on both unconditional and class-conditional video generation tasks, demonstrating consistent improvements across a range of evaluation metrics and datasets.
Our contributions are as follows:
\begin{itemize}
\item We propose a novel metric for assessing both discriminability and temporal consistency of pre-trained vision-encoder features, and use it to systematically analyze their suitability as guidance for video diffusion.
\item We present \framework, a multi-feature fusion and alignment strategy that integrates multiple, but complementary image encoder representations (across frequencies) into video diffusion training.
\item We validate \framework with extensive experiments, demonstrating consistent gains in generation quality and accelerated training on both unconditional and class-conditional video diffusion tasks.
\end{itemize}

\section{Related Work}
\label{sec:background}

\paragraph{Video Diffusion Models.}
Inspired by the success of diffusion models in image generation \cite{rombach2022high,podell2023sdxl}, early approaches in video generation adapted existing image diffusion architectures by incorporating a temporal dimension, enabling models to be trained on both image and video data \cite{blattmann2023align, singer2022make, guo2023animatediff}.
Typically, U-Net-based architectures add temporal attention blocks after spatial attention blocks and expand 2D convolutions to 3D to capture temporal dynamics.
More recently, the introduction of DiT \cite{peebles2023scalable}, combined with large-scale, high-quality curated datasets \cite{ju2025miradata, wang2023internvid, nan2024openvid}, has led to more efficient and scalable diffusion generators \cite{ma2024latte, yang2024cogvideox, zhou2024allegro, videoworldsimulators2024}, significantly advancing video generation quality.
Furthermore, rectified flows \cite{lipman2022flow,esser2024scalingrectifiedflowtransformers, chen2018neural} have shown faster training convergence and inference speed, making them increasingly popular for both image and video generation \cite{esser2024scalingrectifiedflowtransformers, opensora, kong2024hunyuanvideo, hacohen2024ltx}.
Following this wave, we build our method on a video diffusion transformer (V-DiT) architecture and demonstrate its significant efficiency under both diffusion-based and flow-based training paradigms.

\paragraph{Diffusion Training with Auxiliary Loss.}
To enhance diffusion model training, various strategies incorporating auxiliary losses have been proposed. Notably, recent studies have shown that diffusion models inherently learn highly discriminative representations within their features \cite{tang2023emergent, luo2023diffusion, zhang2023tale}. Building on this insight, REPA \cite{yu2024representation} aligns diffusion features with external pretrained visual encoders \cite{oquab2023dinov2, MaskedAutoencoders2021} to improve image diffusion training.
More recently, a range of efforts have been made to enhance video diffusion training.
Track4Gen \cite{jeong2024track4gen} identifies a strong correlation between pixel-space appearance inconsistencies and feature-space temporal inconsistencies—i.e., video point tracking failures—and demonstrates that adding dense point tracking loss to video diffusion features significantly improves generation quality.
Similarly, JOG3R \cite{huang2025unifying} integrates a camera pose estimation loss into video diffusion training, showing that video generation and pose estimation can be jointly optimized.
Additionally, GenRec \cite{weng2025genrec} performs joint training of video generation and video recognition, leveraging mutual reinforcement between these tasks.
VideoJAM \cite{chefer2025videojam} enhances motion coherence by incorporating an optical flow denoising loss, explicitly encoding a strong motion prior into a video diffusion transformer.

Although these works demonstrate that leveraging the auxiliary loss can support the training of video diffusion models, the potential of feature learning with powerful pretrained vision encoders remains underexplored.
In this paper, we investigate whether our proposed additional loss, guided by pretrained vision encoders, eases learning features for video diffusion models.

\paragraph{Self-supervised Vision Encoders.}
Self-supervised learning (SSL) methods have demonstrated remarkable success in vision tasks by learning effective feature representations without human-annotated labels. These methods can be broadly categorized into contrastive learning~\citep{caron2021emerging, oquab2023dinov2}, masked image modeling~\citep{MaskedAutoencoders2021, tong2022videomae}, and predictive modeling~\citep{wang2024dust3rgeometric3dvision, croco, zhang2024monst3r}. 
Many existing studies have not only focused on learning high-quality representations but have also delved into analyzing the features extracted from different vision encoders to gain deeper insights into their underlying structure. Taking a similar approach~\citep{stevens2025sparseautoencodersscientificallyrigorous}, some methods employ a sparse autoencoder to extract interpretable features from vision models. Inspired by this, we perform feature frequency analysis to classify feature characteristics and develop a training framework that enhances generative video diffusion transformer features by leveraging these traits.



\section{\framework}
\label{sec:method}
\begin{figure}[!t]
    \centering
    \includegraphics[width=\columnwidth]{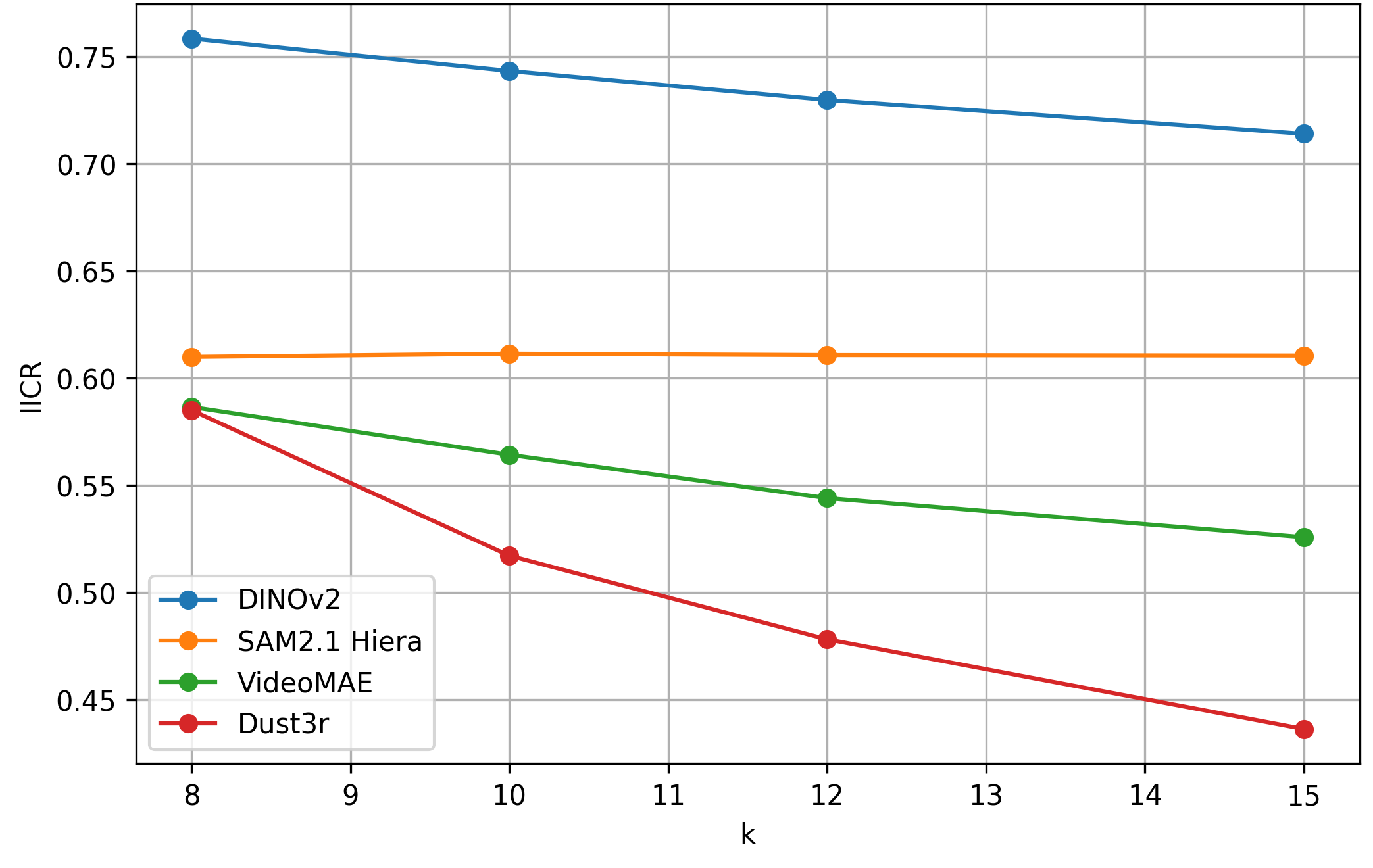} 
    \caption{
    \textbf{IICR comparisons.}
    We present IICR values for different pre-trained vision encoders across varying number of clusters(k) in K-Means over 1,000 videos. DUSt3R and VideoMAE show a decline in IICR as k increases, indicating weaker discriminative power relative to temporal consistency. In contrast, DINOv2 and SAM2 maintain higher values, with DINOv2 slightly decreasing due to its focus on semantics over high-frequency details.}
    \label{fig:iicr_analysis}
\end{figure}

\subsection{Preliminaries}
\label{sec:preliminaries}
\paragraph{Diffusion Models.}
Diffusion models \cite{sohl2015deep, ho2020denoising} generate data by gradually corrupting a clean sample with Gaussian noise and then learning to reverse this noising process. In the forward process, a data sample \(\vx_0\) is progressively corrupted over \(T\) steps. At each time step \(t\), a Gaussian transition is defined as follows:
\begin{equation}
    q(\vx_t \mid \vx_{t-1}) = \mathcal{N}\!\left(\vx_t; \sqrt{1-\beta_t}\,\vx_{t-1},\, \beta_t \mathbf{I}\right)
\label{diffusion_forward}
\end{equation}
where \(\{\beta_t\}\) is a fixed noise scheduler. Using the properties of Gaussian distributions, the marginal distribution \(q(\vx_t \mid \vx_0)\) can be computed in a closed form.

In the reverse diffusion process, the model aims to recover \(\vx_0\) by denoising \(\vx_t\) step by step. Instead of directly predicting \(\vx_{t-1}\), the network \(\boldsymbol{\epsilon}_\theta(\vx_t, t)\) is trained to estimate the noise \(\boldsymbol{\epsilon}\) that was injected during the forward process. The training objective can be expressed as follows:
\begin{equation}
L_{\mathrm{diff}} = \mathbb{E}_{t,\,\vx_0,\,\boldsymbol{\epsilon} \sim \mathcal{N}(\mathbf{0},\mathbf{I})}\!\left[\left\|\boldsymbol{\epsilon} - \boldsymbol{\epsilon}_\theta\!\Bigl(\sqrt{\bar{\alpha}_t}\,\vx_0 + \sqrt{1-\bar{\alpha}_t}\,\boldsymbol{\epsilon},\, t\Bigr)\right\|^2_2\right]
\label{eq:diffusion_loss}
\end{equation}
where $\bar{\alpha}_t=\prod_{s=1}^{t} (1 - \beta_s)$. This formulation of \(\epsilon\) prediction loss has been shown to stabilize training and produce high quality samples.

\paragraph{Flow Models.}
Recently, flow-based training frameworks~\citep{liu2022flow, esser2024scalingrectifiedflowtransformers, chen2025goku, opensora} have gained attention for their ability to progressively transform samples from a prior distribution to the target data distribution through a series of linear interpolations. 
Specifically, we define a velocity field $v_t(\vx)$ of a flow $\psi_t(\vx): [0,1]\times \mathbb{R}^d \rightarrow \mathbb{R}^d$ that satisfies $\psi_t(\x_0)=\x_t$ and  $\psi_1(\x_0)=x_1$.
Here, the $\psi_t$ is uniquely characterized by a flow ODE:
\begin{align}
    d\psi_t(\vx) = v_t(\psi_t(\vx))dt
    \label{eqn:flowode}
\end{align}
%
%
In particular, linear conditional flow  defines the flow as $\x_t=\psi_t(\vx_1|\vx_0) = (1-t)\vx_0 + t \vx_1$. Then, we can compute the velocity field $v_t(\vx_t|\vx_0) = \dot \psi_t(\psi_t^{-1}(\vx_t|\vx_0)|\vx_0)=\vx_1-\vx_0$, leading to the
following conditional flow matching loss:
%
\begin{equation}
\label{eq: flow loss}
L_{\mathrm{RF}} = \mathbb{E}_{t, \, \vx_0,\, \boldsymbol{\epsilon} \sim \mathcal{N}(\mathbf{0},\mathbf{I})} \left[ \left\| (\vx_0 - \boldsymbol{\epsilon}) - v_\theta(\vx_t, t) \right\|^2 \right].
\vspace{-1mm}
\end{equation}
During inference, the predicted velocity is utilized to guide the transformation of an initial noise sample towards the target data distribution through a reverse integration process. This approach enables efficient and controlled generation of high-quality samples.

\begin{figure}[!t] 
    \centering
    \includegraphics[width=\columnwidth]{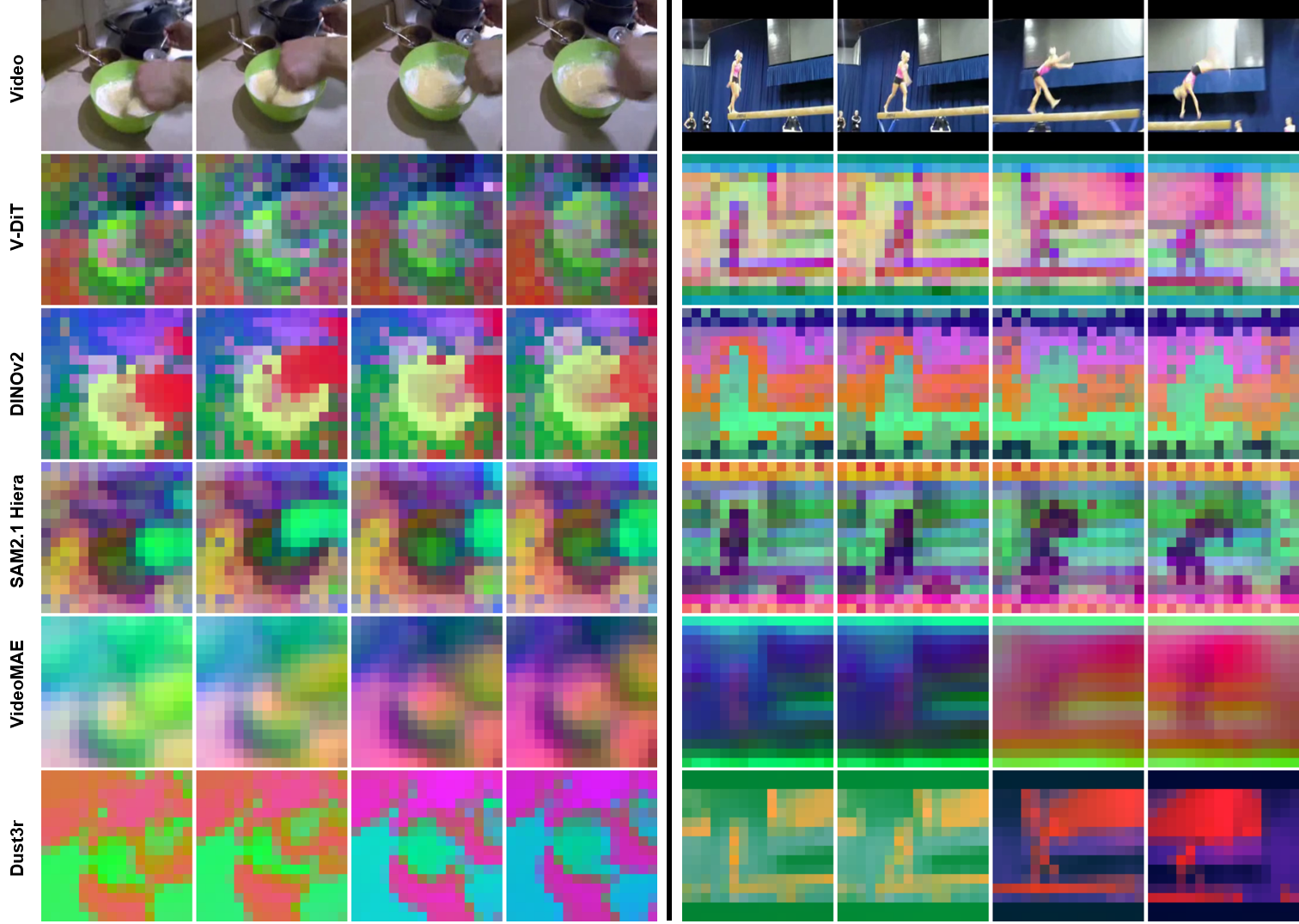} 
    \caption{
    \textbf{PCA visualization of different vision encoder features.} 
    Pretrained image encoders—DINOv2 \cite{oquab2023dinov2} and SAM2 Hiera image encoder \cite{ravi2024sam2}—yield feature representations that remain consistent across frames for both foreground objects and background.
    In contrast, video or 3D vision encoders such as VideoMAE \cite{tong2022videomae} and DUSt3R \cite{wang2024dust3rgeometric3dvision} exhibit significant temporal inconsistencies in their feature representations.
    }
    \label{fig:pca_analysis}
\end{figure}

\subsection{Analysis of Features}
\label{method: feature-analysis}

In order to find optimal vision encoders that accelerate video diffusion model training,
we begin by analyzing the features of the pretrained image, video, and 3D encoders as well as the features of DiTs. 

%
%
%

\paragraph{Discriminative Power and Temporal Consistency.}
An effective pretrained vision encoder for video models should possess two key properties: (1) strong discriminative power and (2) temporal consistency. This is especially crucial in our setting, where the video diffusion model employs a 2D VAE to produce frame-wise independent latents and unpatchifies each frame independently at the final layer of the DiT. In this setup, it is essential that temporally aligned objects yield similar features regardless of their frame positions. To jointly quantify these two properties, we propose a novel metric, \textit{Intra-Inter Consistency Ratio (IICR)}, to evaluate the suitability for video diffusion model training.

Given a video, we extract per-frame feature representations using a pretrained vision encoder. 
Let $\mathbf{u}_i$ be the centroid (mean embedding) of cluster $i$.
We define the inter-cluster distance as
\vspace{-1mm}
\begin{equation}
    D_{\text{inter}} = \min \{ \| \mathbf{u}_i - \mathbf{u}_j \| \mid i \neq j \}.
    \vspace{-1mm}
\end{equation}
which captures the smallest separation between distinct clusters; larger $D_{\text{inter}}$ implies more discriminative features.
We then define the intra-cluster distance $D_{\text{intra}}$ as 
\vspace{-1mm}
\begin{equation}
    D_{\text{intra}} = \max \{ \sigma_i \mid i \text{ in number of clusters} \},
    \vspace{-1mm}
\end{equation}
where $\sigma_i$ is the standard deviation of embeddings within cluster $i$. 
Larger $D_{\text{intra}}$ indicates poorer temporal consistency. 
Finally, we define the Intra-Inter Consistency Ratio (IICR) as
\vspace{-1mm}
\begin{equation}
    \text{IICR} = \frac{D_{\text{inter}}}{D_{\text{intra}}}.
    \vspace{-1mm}
\end{equation}
A higher IICR indicates that the pre-trained encoder produces features that are simultaneously discriminative (large inter-cluster separation) and temporally stable (small intra-cluster variation).

Using the IICR metric, we evaluate four vision encoders: the image-based DINOv2~\cite{oquab2023dinov2} and SAM2.1 Hiera~\cite{ravi2024sam2}, the video encoder VideoMAE~\cite{tong2022videomae}, and DUSt3R~\cite{wang2024dust3rgeometric3dvision}, which is trained for 3D reconstruction.
As illustrated in Fig.~\ref{fig:iicr_analysis}, both DINOv2 and SAM2.1 Hiera consistently achieve high IICR scores across varying $K$.
On the other hand, VideoMAE and DUSt3R degrade noticeably as $K$ increases, indicating a loss in fine-grained discriminability.
We corroborate this with PCA visualizations in Fig.~\ref{fig:pca_analysis}:
the image-level encoders (DINOv2 and SAM2.1 Hiera) produce stable, consistent color assignments for the same backgrounds and subjects across frames, reflecting strong temporal invariance, whereas VideoMAE and the 3D-aware DUSt3R exhibit substantial instability and fluctuations.
These results motivate our selection of DINOv2 and SAM2.1 Hiera as the primary sources for feature alignment in video diffusion.


\paragraph{Frequency Distribution.}

Given the candidate image-level encoders $\mathcal{E}$, we analyze the frequency distribution of their extracted features 
$\mathbf{F} = \mathcal{E}(x_0) \in \mathbb{R}^{h \times w \times c}$. 
We first transform the features into the frequency domain via a 2D Fast Fourier Transform (FFT) and compute the log-magnitude spectrum:
\begin{equation}
\mathbf{S} = \mathrm{FFT}_{\mathrm{2D}}(\mathbf{F}),
\end{equation}
\begin{equation}
\mathbf{M} = \log \left( |\mathbf{S}| + \epsilon \right),
\end{equation}
where $\epsilon$ is a small constant for numerical stability.  

We then measure the mean log-magnitude difference between the lowest- and highest-frequency regions.  
The lowest-frequency (DC) component is given by:
\begin{align}
\text{Lowest frequency:} \quad & \mathbf{M}_{0,0},
\end{align}
where $(0,0)$ denotes the center of the Fourier spectrum.  
The high-frequency (HF) region is defined as:
\begin{align}
\text{High-frequency region:} \quad 
& \mathcal{H} = \left\{ (u,v) \,\middle|\, \sqrt{u^2 + v^2} > 0.75\,r \right\},
\end{align}
where $r = h/2$ for a square input of size $h \times h$.  

The frequency difference metric is computed as
\begin{equation}
\Delta_{\mathrm{freq}} 
= \frac{1}{|\mathcal{H}|} \sum_{(u,v) \in \mathcal{H}} \mathbf{M}_{u,v} 
- \mathbf{M}_{0,0}.
\end{equation}

\begin{figure}[t]
    \centering
    \includegraphics[width=\columnwidth]{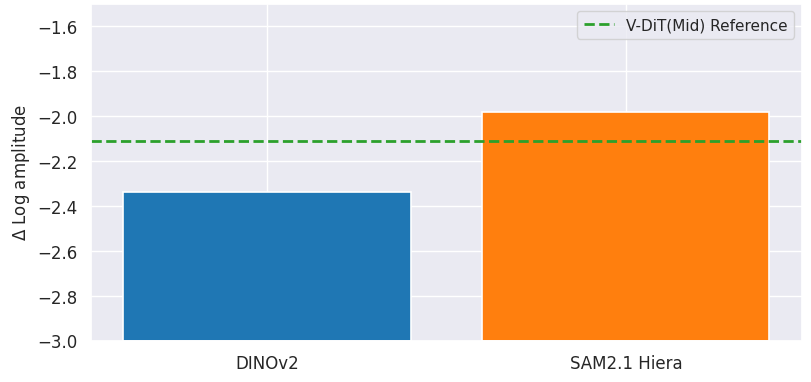} 
    \caption{
    \textbf{Mean log differences between the lowest and highest frequency components.} DINOv2 focuses more on low-frequency components, whereas SAM2.1 Hiera places greater emphasis on high-frequency regions. We also plot the same metric for the features obtained from a mid-block of a video diffusion transformer model as a reference.}
    \label{fig:freq_analysis}
\end{figure}

As shown in Fig.~\ref{fig:freq_analysis}, the value $\Delta_{\mathrm{freq}}$ for DINOv2 and SAM2.1 Hiera differs by $0.3$ in log scale, corresponding to a ${\times}1.35$ ratio.  
This indicates a clear gap in their frequency composition: DINOv2~\citep{oquab2023dinov2} predominantly captures low-frequency, semantic structures, while the Hiera~\citep{ryali2023hiera} encoder in SAM2.1~\citep{ravi2024sam2} exhibits stronger sensitivity to high-frequency details.  
Given the complementary nature of these features, we incorporate both into our video diffusion training, as discussed in the following section.

%

\subsection{Video DiT Training with Feature Alignment}
\label{method: training}

Guided by the prior analysis, we design a training framework that enforces consistency between the intermediate tokens of a video diffusion transformer (V-DiT) and representations from pre-trained image encoders.

\paragraph{Patch Token Alignment.} 
At the heart of our method is \emph{patch token alignment}, which regularizes V-DiT by aligning its patch-level tokens with those extracted from a pre-trained image transformer $T_I$.
For each video frame, $T_I$ produces local patch embeddings $p_i$, and V-DiT produces corresponding tokens $q_i$, where $i$ denotes the patch or token index. 
To bridge the gap between their feature spaces, we pass $q_i$ through a lightweight mapper—implemented as a multi-layer perceptron (MLP)—before alignment.
Feature alignment is then enforced by minimizing:
\begin{align}
    \mathcal{L}_{\text{align}} = \frac{1}{N}\sum_{i=1}^{N} \ell(p_i, \text{MLP}(q_i))
    \label{eq:align}
\end{align}
where $\ell(\cdot,\cdot)$ is a distance measure (e.g., cosine distance) and $N$ is the total number of patch tokens. 
The full training objective augments the base video denoising loss (diffusion or flow-based) with the alignment term:
\begin{equation}
    \mathcal{L}_{\text{total}} = \mathcal{L}_{\text{diff/RF}} + \gamma \cdot \mathcal{L}_{\text{align}}
    \label{eq:total_loss}
\end{equation}
where $\gamma$ balances the contribution of the alignment regularizer.

\paragraph{Multi-Feature Fusion.} 
We further enhance training by aligning V-DiT features with a fused representation constructed from multiple complementary image encoders.
Different encoders capture distinct frequency characteristics—e.g., DINOv2 emphasizes low-frequency semantics, while SAM2.1 Hiera captures high-frequency detail—so we combine their strengths.
Given a set of encoder outputs $F_1, F_2, \dots, F_n$, we construct a unified feature representation via concatenation along the feature dimension:
\begin{equation}
    F = [F_1; F_2; \dots; F_n]
\end{equation}
producing a richer, multi-frequency supervisory signal.
This aggregated representation enforces consistency across diverse feature sources and preserves both coarse and fine-grained information. 
In specific, we integrate DINOv2 and SAM2.1 Hiera, leveraging their complementary strengths—DINOv2 captures low-frequency semantics, while SAM2.1 Hiera focuses on high-frequency details. 
This combination enables a more holistic representation enables capturing nuanced details beyond mere high-level semantics.

\section{Experiments}
\label{sec:result}
\subsection{Experiment Setup}

\paragraph{Implementation details.}
We validate the performance of \framework in both unconditional and class-conditional video generation tasks.
For class-conditional video generation, we implement the V-DiT model at two scales, V-DiT-L and V-DiT-XL, following previous practice \cite{ma2024latte}. The patch size of 2 is utilized for all V-DiT models.
V-DiT-L consists of $24$ transformer blocks with a total of 466M parameters in its transformer backbone, while V-DiT-XL comprises $28$ transformer blocks with 683M parameters.
We use the UCF-101 \cite{soomro2012ucf101dataset101human}, SkyTimelapse \cite{xiong2018learning}, and FaceForensics \cite{rossler2018faceforensics} datasets for training, with videos at a uniform resolution of $256 \times 256 \times 16$.
We train V-DiT-L for 1M steps with a batch size of 24 and V-DiT-XL for 200K steps with a batch size of 16.
For video sampling after training, we use the DDIM sampler~\citep{song2020denoising} with 50 steps for diffusion-based models and the Euler sampler with 50 steps for flow-based models.

\paragraph{Baselines.}
We compare our method (denoted as \emph{`V-DiT + Ours'}) to several baselines. First, we train the V-DiT-L and V-DiT-XL models using only the standard diffusion or flow loss which we denote as \emph{`V-DiT'}.
Next, we also train the V-DiT models using only a single pretrained image encoder for feature alignment, either DINOv2 or SAM2.1 Hiera.

\paragraph{Evaluation metrics.}
We assess video generation quality using Fréchet Video Distance (FVD)~\citep{unterthiner2019accurategenerativemodelsvideo}, Fréchet Inception Distance (FID)~\citep{heusel2018ganstrainedtimescaleupdate}, Inception Score (IS)~, and frame-wise CLIP similarity~\citep{Radford2021LearningTV}. 
Classical FVD computed with the I3D backbone tends to emphasize per-frame fidelity over motion. 
To counterbalance this bias and better capture both content and temporal dynamics, we adopt the content-debiased FVD variant~\citep{ge2024content} using VideoMAE~\citep{tong2022videomae} as the feature extractor.

\begin{figure*}[t] 
    \centering
    \includegraphics[width=1.0\textwidth]{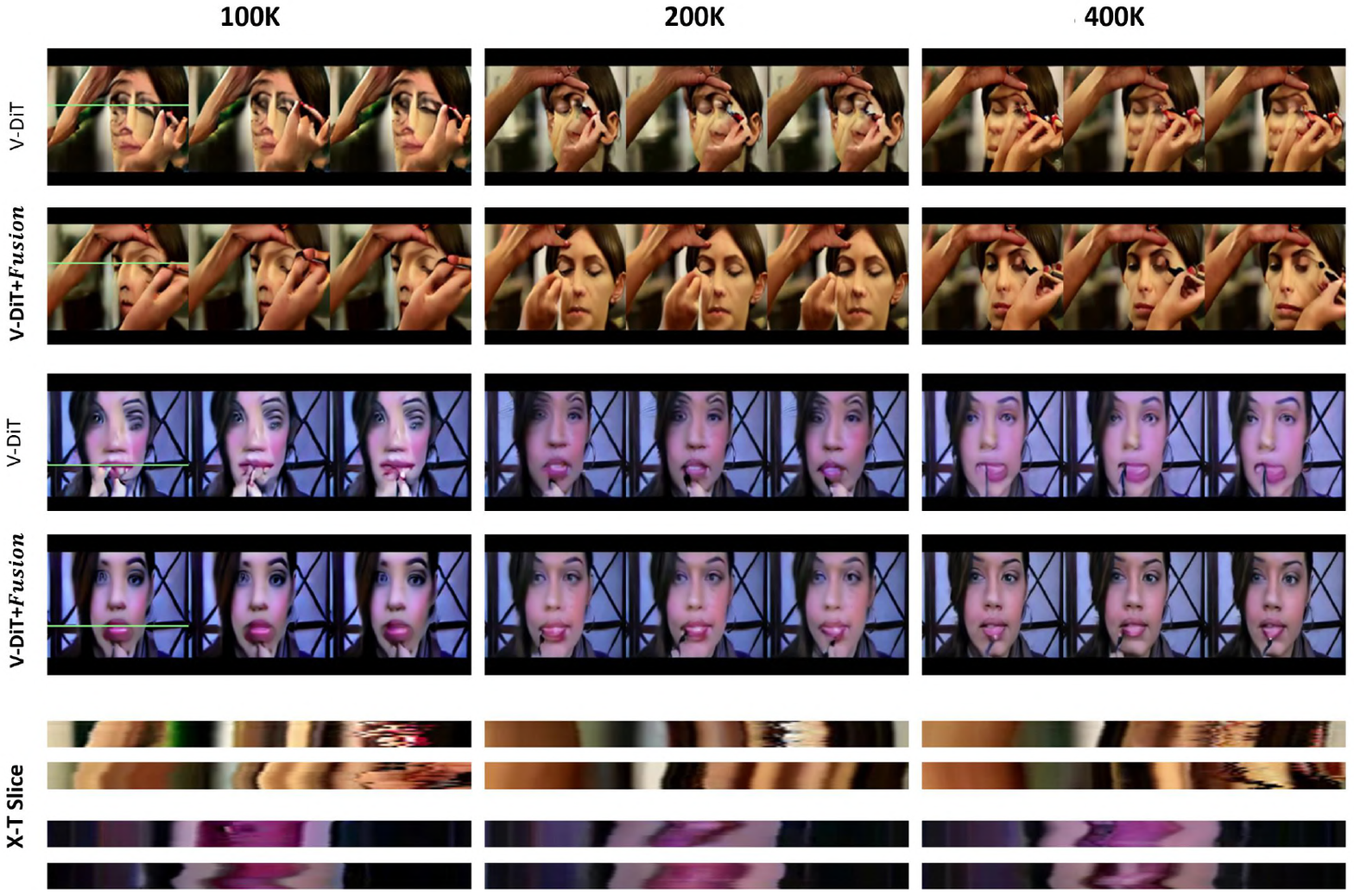} 
    \caption{
    \textbf{Qualitative comparisons on UCF-101}. Our method converges noticeably faster than the original model. Moreover, in addition to improved per-frame image quality, the X-T slice analysis reveals smoother frame transitions, highlighting enhanced visual fidelity and motion consistency in the V-DiT model. Here, the green line indicates the position of the X-T slice, and videos in the same row share the same green line.}
    \label{fig:main_fig}
\end{figure*}

\subsection{Results}

\paragraph{Class-conditional Video Generation.} 
Tab.~\ref{tab:main} demonstrates that \framework yields substantial gains in both FVD \cite{unterthiner2019accurategenerativemodelsvideo} and FID \cite{heusel2018ganstrainedtimescaleupdate} over the vanilla V-DiT baseline, reflecting improvements in overall video fidelity as well as per-frame image quality.
Moreover, aligning with a single image encoder (V-DiT + Ours (DINOv2) or V-DiT + Ours (SAM2)) already outperforms the original model, while fusing DINOv2 and SAM2.1 Hiera (V-DiT + Ours (Fusion) delivers the best FVD and FID.
Notably, the fusion variant at 400K steps surpasses the baseline at 1M steps, indicating at least a $\times 2.5$ faster convergence, substantially reducing the computational cost of video diffusion model training.

Feature alignment does not compromise diversity: the Inception Score increases, implying quality gains are not due to mode collapse.
For temporal consistency (measured via frame-wise CLIP similarity), \framework consistently matches or exceeds the vanilla baseline, with differences among non-baseline variants being negligible—showing maintained scene-level coherence.

Qualitatively (Fig.~\ref{fig:main_fig}), V-DiT + Ours (Fusion) produces sharper and more anatomically coherent human structures, and X–T slice visualizations reveal smoother motion compared to the baseline. 

\begin{table}[h]
    \centering
    \begin{adjustbox}{max width=\columnwidth}
    \begin{tabular}{lcccccc}
        \toprule
        Method & Iteration & FVD $\downarrow$ & FID $\downarrow$ & IS $\uparrow$ & CLIP $\uparrow$ \\
        \midrule
        V-DiT & 400K & 262.77 & 43.87 & 78.23$\pm$1.65 & 0.9309 \\
        V-DiT + Ours (DINOv2) & 400K & 225.32 & 40.88 & \underline{82.46$\pm$1.53} & \underline{0.9329} \\
        V-DiT + Ours (SAM2) & 400K & \underline{220.16} & \underline{40.65} & 81.99$\pm$1.54 & 0.9321 \\
        V-DiT + Ours (\emph{Fusion}) & 400K & \textbf{206.73} & \textbf{38.89} & \textbf{83.16$\pm$1.52} & \textbf{0.9333} \\
        
        \midrule
        V-DiT & 1M & 221.63 & 40.70 & 81.62$\pm$1.57 & 0.9331 \\
        V-DiT + Ours (DINOv2) & 1M & \underline{194.70} & \underline{38.19} & \underline{84.56$\pm$1.48} & \textbf{0.9340} \\
        V-DiT + Ours (SAM2) & 1M & 196.48  & 38.80 & 83.54$\pm$1.50 & \underline{0.9338} \\
        V-DiT + Ours (\emph{Fusion}) & 1M & \textbf{187.46} & \textbf{37.31} & \textbf{84.67$\pm$1.49} & \underline{0.9338} \\
        \bottomrule
    \end{tabular}
    \end{adjustbox}
    \caption{
    \textbf{Quantitative comparisons of V-DiT-L training on UCF-101 (Class-conditional generation)}.
    The Fusion method generates higher-quality videos more than 2.5 times faster than Vanilla and achieves the best performance even after extended training. 
    While using a single feature alignment significantly improves quality, the Fusion method further amplifies this effect, maximizing the benefits of alignment. 
    Here, SAM2 refers to the SAM2.1 Hiera image encoder. 
    }

    \label{tab:main}
\end{table}

\begin{table}[h]
    \centering
    \begin{adjustbox}{max width=0.9\columnwidth}
    \begin{tabular}{clcccc}
        \toprule
        Dataset & Method & Iteration & FVD $\downarrow$ & FID $\downarrow$ & CLIP $\uparrow$ \\
        \midrule
        \multirow{4}{*}{SkyTimelapse} 
        & V-DiT & 50K & 250.54  & 50.44 & 0.9684 \\
        & V-DiT + Ours (\emph{Fusion}) & 50K & 245.42 & 49.86 & 0.9688 \\
        & V-DiT & 200K & 200.51 & 39.54 &  0.9723 \\
        & V-DiT + Ours (\emph{Fusion}) & 200K & \textbf{196.01} & \textbf{38.38} & \textbf{0.9724} \\
        \midrule
        \multirow{4}{*}{FaceForensics} 
        & V-DiT & 50K & 91.59 & 18.28 & 0.9466 \\
        & V-DiT + Ours (\emph{Fusion})& 50K & 76.82 & 16.16 & 0.9470 \\
        & V-DiT & 200K & 45.77 & 6.02 &  0.9514 \\
        & V-DiT + Ours (\emph{Fusion}) & 200K & \textbf{39.86} & \textbf{5.60} & \textbf{0.9526} \\
        \bottomrule
    \end{tabular}
    \end{adjustbox}
    \caption{
    \textbf{Quantitative comparisons of V-DiT-L training on SkyTimelapse and FaceForensics (Unconditional generation).}
    Our method consistently improves upon the baseline model, with a particularly notable increase in performance for FaceForensics. We use a depth of 8 for feature alignment.
    }
    \label{tab:main_2}
\end{table}

\paragraph{Unconditional Video Generation.}

\begin{figure}[t] 
    \centering
    \includegraphics[width=1.0\columnwidth]{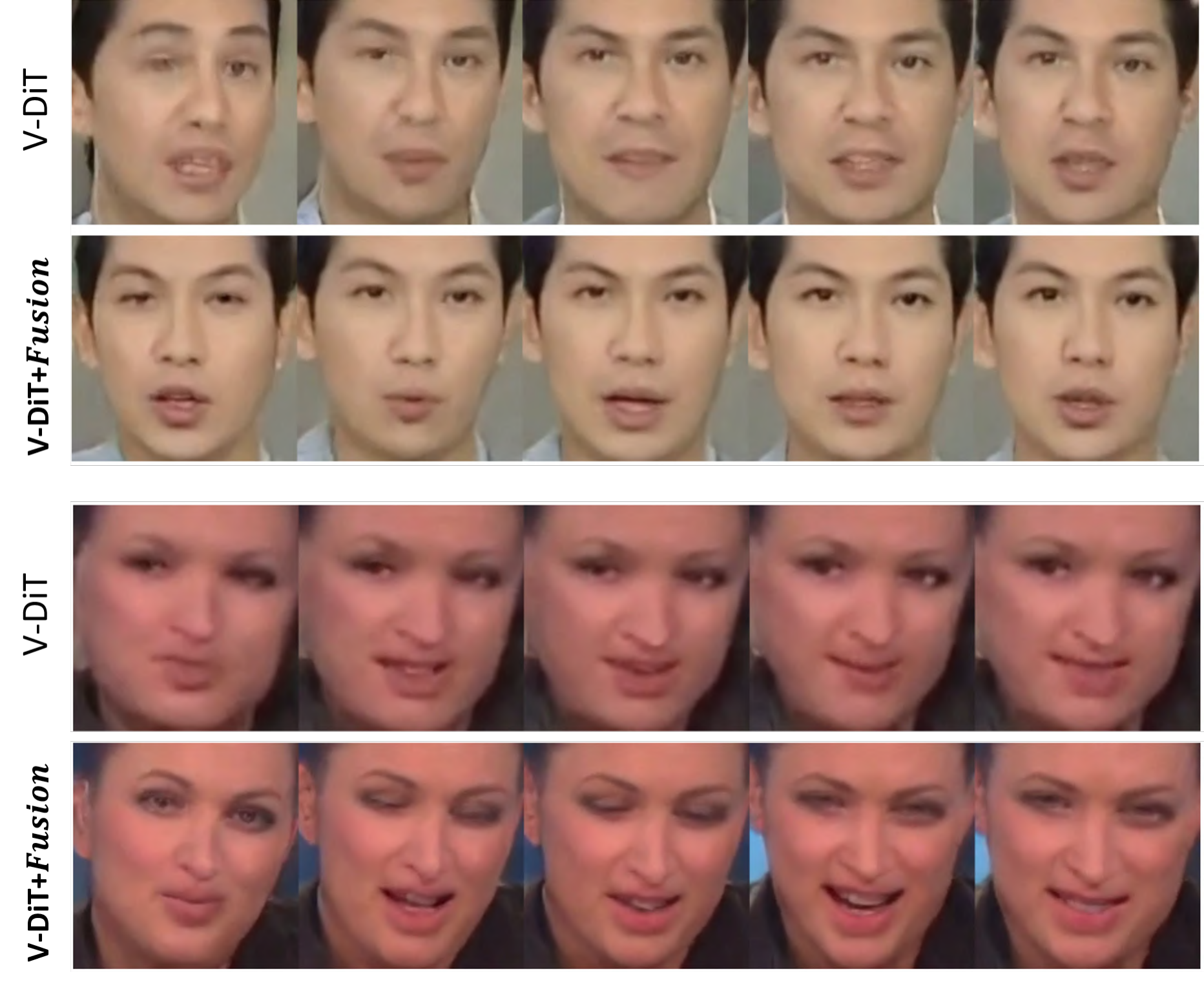} 
    \caption{
    \textbf{Qualitative comparison on FaceForensics}. 
    Our Fusion method preserves finer details and achieves higher fidelity than the baseline}
    \label{fig:main_ffs}
\end{figure}

As shown in Tab.~\ref{tab:main_2}, our method consistently outperforms the baseline across all metrics. However, in the case of the SkyTimeLapse dataset, the performance gap is notably small. We attribute this to the characteristics of the dataset itself. Unlike other datasets~\citep{soomro2012ucf101dataset101human}, SkyTimeLapse primarily captures sky scenes with minimal object presence. As a result, the discriminability of pre-trained vision encoder plays a limited role in this setting. It means that for datasets where discriminability is minimal or unnecessary, the role of feature alignment is significantly reduced. This will undoubtedly be an important insight for future video diffusion model training.

In contrast, for subject-centric datasets such as FaceForensics, the effectiveness of our vision encoder's discriminability is more pronounced, leading to significant performance improvements as we can see in Tab.~\ref{tab:main_2}. As seen in Fig.~\ref{fig:main_ffs}, at the same training iteration step 50K, our method produces videos with higher fidelity, where the shape and position of human faces are more clearly defined. 

\subsection{Ablation Studies}
In this section, we provide a thorough analysis of the various design choices we make in \framework.

\vspace{-2pt}
\paragraph{Effect of denoising objectives for V-DiT.}
We integrate \framework with two prevalent denoisingV-DiT objectives: flow loss $\mathcal{L}_{\text{RF}}$ and diffusion loss $\mathcal{L}_{\text{diff}}$.
As shown in Tab~\ref{tab:ablation_study diffusion vs flow loss}, \framework yields substantial FVD improvements for V-DiT-XL under both setups, with the fusion variant delivering the best overall results.
This confirms that \framework consistently accelerates training and boosts quality independent of the underlying objective.

\vspace{-2pt}

\paragraph{Effect of feature alignment layer.}
An important design choice in \framework is which transformer block to use in V-DiT to perform feature alignment. Hence, we conducted experiments using blocks at different depths and observed performance variations across models, as shown in Table~\ref{tab:ablation_study alignment layer}.
For the V-DiT-XL model, aligning at a mid-level depth produced slightly improved results. This suggests that the $depth = 12$ performs better as it has a sufficient number of blocks to effectively capture semantic representations.
Additionally, aligning features immediately after the spatial block—rather than after the temporal block—yielded a slight performance gains.
The difference is minimal, but the temporal block only considers embeddings at the same position rather than neighboring pixels, making it more challenging to directly model discriminative features.

These insights reinforce the robustness of our methodology across different configurations and emphasize the strategic advantage of spatially guided feature alignment for improving both training efficiency and final model quality.

\begin{table}[t]
    \centering
    \begin{adjustbox}{max width=\columnwidth}
    \begin{tabular}{ccccc}
        \toprule
         \multirow{2}{*}{Method} & \multirow{2}{*}{Depth} & \multicolumn{2}{c}{Alignment} & \multirow{2}{*}{FVD} \\
        \cmidrule(lr){3-4}
         &  & Spatial & Temporal & \\
        \midrule 
         V-DiT & 12 & - & - & 379.90 \\
         \hdashline
         + Ours (DINOv2) & 12 & \checkmark & - & \textbf{311.14} \\
         + Ours (DINOv2) & 8 & \checkmark & - & \underline{313.18} \\
         + Ours (DINOv2) & 12 & - & \checkmark & 315.41 \\
        \bottomrule
    \end{tabular}
    \end{adjustbox}
    \caption{
    \textbf{Quantitative ablation on feature alignment layer.}
    We train the V-DiT-XL model on the UCF-101 dataset \cite{soomro2012ucf101dataset101human} using joint diffusion and alignment loss for 200K iterations, exploring the effects of spatial, temporal, and depth variations.}
    \label{tab:ablation_study alignment layer}
\end{table}

\begin{table}[t]
    \centering
    \begin{adjustbox}{max width=\columnwidth}
    \begin{tabular}{c|cccccc}
        \toprule
        Features & DINOv2 & SAM2 & VideoMAE & DUSt3R & SD3 \\
        \midrule
        FVD & \textbf{301.67} & \underline{314.12} & 323.11 & 384.17 & 351.03 \\
        \bottomrule
    \end{tabular}
    \end{adjustbox}
    \caption{
    \textbf{Quantitative ablation on using different feature representations.}
    For this experiment, we train on UCF-101 dataset \cite{soomro2012ucf101dataset101human} with joint flow and alignment loss for 200K iterations.
    For the alignment, spatial layer at depth 12 of V-DiT-XL is uniformly used.
    }
    \label{tab:ablation_study different encoders}
\end{table}

\begin{table}[t]
    \centering
    \begin{adjustbox}{max width=\columnwidth}
    \begin{tabular}{c|cc}
        \toprule
        Integration Method & 2-MLP & Fusion \\
        \midrule
        FVD & 307.63 & \textbf{295.63} \\
        \bottomrule
    \end{tabular}
    \end{adjustbox}
    \caption{
    \textbf{Quantitative ablation on using two separate MLPs versus feature concatenation.}
    For this experiment, we train on UCF-101 dataset \cite{soomro2012ucf101dataset101human} with joint flow and alignment loss for 200K iterations.
    For the feature alignment, we use DINOv2 and SAM2.1 (Hiera) features and spatial layer at depth 12 of V-DiT-XL is uniformly used.
    }
    \label{tab:ablation_study mlp vs fusion}
\end{table}

\begin{table}[t]
    \centering
    \begin{adjustbox}{max width=0.7\columnwidth}
    \begin{tabular}{clc}
        \toprule
         \multirow{2}{*}{Objective} & \multirow{2}{*}{Method} & \multirow{2}{*}{FVD} \\
          &  & \\
        \midrule 
         \multirow{4}{*}{Diffusion Loss}  
         & V-DiT & 379.90 \\
         &  + Ours (DINOv2) & 311.14 \\
         & + Ours (SAM2)  & 314.13 \\
         & + Ours (\emph{Fusion}) & \underline{298.76} \\
         \cmidrule(lr){1-3}
         \multirow{4}{*}{Flow Loss}  
         & V-DiT & 369.30 \\
         & + Ours (DINOv2) & 301.67 \\
         & + Ours (SAM2) & 314.12 \\
         & + Ours (\emph{Fusion}) & \textbf{295.63} \\
        \bottomrule
    \end{tabular}
    \end{adjustbox}
    \caption{
    \textbf{Quantitative ablation on denoising objective: diffusion loss vs flow loss.}
    We train V-DiT-XL on UCF-101 dataset \cite{soomro2012ucf101dataset101human} for 200K iterations.
    For the feature alignment, spatial layer at depth 12 is uniformly used.
    }
    \label{tab:ablation_study diffusion vs flow loss}
\end{table}

\vspace{-3pt}

\paragraph{Effect of pretrained feature representations.}
We have used DINOv2 and SAM2.1 Hiera features in \framework based on our prior analysis of different feature representations with respect to discriminative power and temporal consistency. In order to further validate this choice, we perform V-DiT training using different pretrained feature representations including DUSt3R, VideoMAE, and additionally Stable Diffusion 3. As illustrated in Tab~\ref{tab:ablation_study different encoders}, DUSt3R leads to degraded performance which we attribute to temporal inconsistencies it suffers from. Use of VideoMAE also yields suboptimal results compared to using image encoders.
%
We attribute the primary reason for VideoMAE’s underperformance to its temporal compression mechanism, which reduces the temporal dimension by a factor of two. This compression can disrupt temporal consistency in the extracted features, leading to instability during V-DiT training. Based on these results, we observed that the performance metrics follow the order defined by our proposed IICR metric. Therefore, when training a video diffusion model, IICR can be used to estimate the expected performance without the need to experimentally evaluate all features directly.

Additionally, we test the performance of using pre-trained features from Stable Diffusion 3 (SD3)~\citep{esser2024scalingrectifiedflowtransformers} (discussion of which feature we use is in Suppl.~\ref{suppl:pca-features}). While SD3 contributes to a moderate improvement in performance, its computationally intensive nature renders it an inefficient choice for practical applications. The high computational cost associated with SD3 diminishes its viability, particularly when aiming for a more efficient and scalable video diffusion model.

\vspace{-5pt}

\paragraph{Effect of multi-feature concatenation.}
\framework concatenates multiple feature representations as a fusion mechanism. An alternative approach is to use two separate MLPs to map the features of the video diffusion model into the feature spaces of two different pre-trained vision encoders. 
Tab.~\ref{tab:ablation_study mlp vs fusion} shows that the fusion approach significantly outperforms the method of using separate mapper MLPs. 
This indicates that leveraging a unified feature space leads to more effective representation learning and improved performance.

\section{Conclusion}
\label{sec:conclusion}

In this work, we introduce \framework, a framework designed to reduce the training cost of video diffusion models by incorporating key insights and methodologies. Specifically, we propose a new metric to evaluate the suitability of various pretrained vision encoders for video diffusion model training. Our results show that performance trends follow a similar pattern based on this metric. Additionally, we leverage pretrained image encoder features that preserve temporal consistency and propose a novel feature fusion approach that exploits a broad range of frequency bands. By aligning these enhanced features within the Video DiT transformer block, we not only improve video diffusion model training but also accelerate convergence. We validate our method on un-conditional and class-conditional diffusion models, demonstrating its effectiveness in reducing training costs. We anticipate that our approach will play a significant role in advancing cost-efficient training strategies for future video diffusion models.

\section{Acknowledgement}
\label{sec:acknowledgement}

We are deeply grateful to Sihyun Yu for providing constructive feedback and engaging in helpful discussions throughout the development of this work. The thoughtful comments and valuable suggestions offered at various stages have been instrumental in refining our methodology.

\bibliography{aaai2026}

\clearpage
\appendix
\setcounter{page}{1}

\noindent This supplementary material is organized as follows:
\begin{enumerate}[label=(\Alph*)]
    \item Evaluation Metrics
    \item Additional Experimental Details
    \item V-DiT Architecture
    \item Additional Feature Analysis
    \item Limitation
    \item Comparison to Previous Methods
    \item Additional Qualitative Results
\end{enumerate}

\section{Evaluation Metrics}
\label{suppl:evaluation}

\paragraph{Content-debiased FVD.}
In videos, when image quality is similar but motion appears unrealistic, FVD often produces overly favorable scores. Since motion quality is crucial for video evaluation, computing FVD using features extracted from a large-scale unsupervised model instead of the conventional I3D network can mitigate this bias, leading to a more reliable assessment ~\citep{ge2024content}. Therefore, we extracted features using the VideoMAE model and computed FVD based on those features.
\paragraph{FVD Protocol.} To compute the Fréchet Video Distance (FVD), we follow the evaluation protocol of StyleGAN-V~\citep{stylegan_v}. Specifically, we use the UCF-101 dataset, where we first split the dataset into 9,537 training and 3,783 testing subsets. The model is trained exclusively on the training set, while evaluation is performed on a subset of the test set.

For evaluation, we sample 2,048 videos from the test split, ensuring that the number of selected videos per class maintains the original class distribution within the test set. Within each class, videos are randomly selected according to its proportion. From each selected video, we extract a single 16-frame clip at a random temporal location, ensuring that each video contributes only one clip to the final evaluation set. These extracted clips are then processed through the VideoMAE network to obtain content-debiased FVD features, which are used to compute the real feature statistics.

For the generated (fake) feature statistics, we generate 2,048 videos \textbf{without} classifier-free guidance and process them in the same manner as the real videos. That is, a single random 16-frame clip is extracted from each generated video, and feature statistics are computed.

The Fréchet Video Distance (FVD) is then computed using the following formulation:
\begin{equation}
    \text{FVD} = \left\| \mu_r - \mu_g \right\|^2 + \text{Tr} \left( \Sigma_r + \Sigma_g - 2 (\Sigma_r \Sigma_g)^{\frac{1}{2}} \right)
\end{equation}
where:
\begin{itemize}
    \item \( \mu_r \) and \( \Sigma_r \) denote the mean and covariance matrix of the real feature distribution.
    \item \( \mu_g \) and \( \Sigma_g \) denote the mean and covariance matrix of the generated (fake) feature distribution.
    \item \( \text{Tr}(\cdot) \) represents the trace of a matrix.
\end{itemize}

This metric quantifies the distributional discrepancy between real and generated video feature embeddings, where lower FVD values indicate higher fidelity and temporal coherence in the generated videos.

\paragraph{FID Protocol.} 
To compute FID for video generators, we follow the approach used in StyleGAN-V. Specifically, we generate 2,048 videos, each containing 16 frames, and use all frames for FID computation. For real statistics, similar to the FVD protocol, we sample 2,048 clips from these videos with a random starting frame \( t \). This results in approximately 33,000 images for constructing the dataset. However, since consecutive frames within a video share substantial content, the dataset exhibits lower diversity compared to the standard 50,000-image set typically used for evaluating image generators. While generating 50,000 videos and sampling a single frame from each would provide a more comprehensive assessment of image quality, the computational cost of such an approach is prohibitively high.

\paragraph{IS.}
We compute the Inception Score (IS) using a C3D~\citep{tran2015learningspatiotemporalfeatures3d} network fine-tuned on the UCF-101 dataset. To assess diversity, we calculate the IS score using 2,048 generated videos. The IS score is computed as follows:
\begin{equation}
IS = \exp \left( \mathbb{E}_{\mathbf{x}} \left[ D_{KL} \left( p(y | \mathbf{x}) \parallel p(y) \right) \right] \right),
\end{equation}
where $p(y | \mathbf{x})$ is the predicted class distribution of the video $\mathbf{x}$, and $p(y)$ is the marginal distribution obtained by averaging $p(y | \mathbf{x})$

\paragraph{Frame-wise CLIP similarity.}
To evaluate frame-wise clip similarity, we compute the cosine similarity between the CLIP features of the first frame and each preceding frame within a video. This metric helps assess whether significant changes occur in the overall background or between consecutive frames. Specifically, for a video with $T$ frames, we compute the similarity between the first frame $\mathbf{x}_1$ and each subsequent frame $\mathbf{x}_t$ ($t = 2, \dots, T$), and take the average over all $T-1$ comparisons:
\begin{equation}
S_{\text{clip}} = \frac{1}{T-1} \sum_{t=2}^{T} \frac{1}{2}[\cos \left( \phi(\mathbf{x}_1), \phi(\mathbf{x}_t) \right) + \cos(\phi(\mathbf{x}_{t-1}),\phi(\mathbf{x}_t))]
\end{equation}
where $\phi(\mathbf{x})$ represents the CLIP feature embedding of frame $\mathbf{x}$, and $\cos(\cdot, \cdot)$ denotes the cosine similarity. A higher score indicates greater consistency in the background and smoother transitions between frames.


\begin{figure}[t]
    \centering
    \newcommand{\numColumns}{4}
    \newcommand{\columnSpacing}{0.1cm}

    \begin{tabular}{
        @{}
        p{\dimexpr(\columnwidth-\columnSpacing*(\numColumns-1))/\numColumns} 
        @{\hspace{\columnSpacing}}
        p{\dimexpr(\columnwidth-\columnSpacing*(\numColumns-1))/\numColumns} 
        @{\hspace{\columnSpacing}}
        p{\dimexpr(\columnwidth-\columnSpacing*(\numColumns-1))/\numColumns} 
        @{\hspace{\columnSpacing}}
        p{\dimexpr(\columnwidth-\columnSpacing*(\numColumns-1))/\numColumns} 
        @{}
    }
        \animategraphics[loop, width=\linewidth]{8}{rebuttal_videos/image_pca/}{00}{15} &
        \animategraphics[loop, width=\linewidth]{8}{rebuttal_videos/image_frames/}{00}{15} &
        \animategraphics[loop, width=\linewidth]{8}{rebuttal_videos/video_pca/}{00}{15} &
        \animategraphics[loop, width=\linewidth]{8}{rebuttal_videos/video_frames/}{00}{15}
    \end{tabular}

    \vspace{-4pt}
    \begin{tabularx}{\columnwidth}{XXXX}
        \centering \tiny (a) SAM2.1 PCA&
        \centering \tiny (b) SAM2.1 Track&
        \centering \tiny (c) VideoMAE PCA&
        \centering \tiny (d) VideoMAE Track
    \end{tabularx}
    \caption{ \footnotesize
   PCA visualization and tracking comparison videos of image and video encoders. \textit{Click each image to play the video in Acrobat Reader.}
    }
    \label{fig:tracking}
\end{figure}

\section{Additional Experimental Details}
\label{suppl:experimental-details}

\paragraph{Training and Inference using \framework.}
For patch token alignment, a single MLP is employed to map all patches, rather than using separate MLPs for each patch. This ensures a consistent mapping across patches through a shared MLP. Additionally, when performing concatenation, the feature vectors should first be normalized along the feature dimension before concatenation. This normalization step is crucial because the projection loss is computed using cosine similarity. Without normalization, discrepancies in the scale of the feature vectors could cause one feature to dominate the other, thereby diminishing its influence. By normalizing both feature vectors before concatenation, we ensure a balanced alignment. During training, this alignment process is applied, while at inference, the MLP is removed, and the video diffusion model is used as in standard inference.

\begin{algorithm}[H]
\caption{Training Procedure}
\label{alg:training}
\begin{algorithmic}[1]
\Require Video $\mathcal{V}$, VAE $\mathcal{E}_{\text{VAE}}$, DINOv2 $\mathcal{E}_{\text{DINOv2}}$, SAM2.1 Hiera $\mathcal{E}_{\text{SAM}}$, trainable $MLP(\cdot)$, and $\gamma$.
\Ensure Total Loss $\mathcal{L}_{\text{total}}$

\State \textbf{// Step 1: Encode video using VAE}
\State $\mathbf{z} \gets \mathcal{E}_{\text{VAE}}(\mathcal{V})$ \Comment{Encode the video $\mathcal{V}$}

\State \textbf{// Step 2: Extract features using Vision Encoders}
\State $F_1 \gets \mathcal{E}_{\text{DINOv2}}(\mathcal{V})$ \Comment{Extract feature $F_1$ using $\mathcal{E}_{\text{DINOv2}}$}
\State $F_2 \gets \mathcal{E}_{\text{SAM}}(\mathcal{V})$ \Comment{Extract feature $F_2$ using $\mathcal{E}_{\text{SAM}}$}

\State \textbf{// Step 3: Normalize and concatenate features}
\State $F \gets \Bigl[\frac{F_1}{\|F_1\|};\ \frac{F_2}{\|F_2\|}\Bigr]$ \Comment{Feature Fusion}

\State \textbf{// Step 4: Compute Projection Loss}
\State $f \gets MLP(\text{intermediate diffusion feature})$
\State $\mathcal{L}_{\text{proj}} \gets 1 - \cos\bigl(f,\, F\bigr)$ 

\State \textbf{// Step 5: Calculate Total Loss}
\State $\mathcal{L}_{\text{total}} \gets \mathcal{L}_{\text{diffusion}} + \gamma\, \mathcal{L}_{\text{proj}}$ 

\State \Return $\mathcal{L}_{\text{total}}$
\end{algorithmic}
\end{algorithm}

\begin{algorithm}[H]

\caption{Inference Procedure}
\label{alg:inference}
\begin{algorithmic}[1]
\Require Flow model \(f(\mathbf{z}, t)\), initial latent \(\mathbf{z}_T\), VAE Decoder \(\mathcal{D}_{\text{VAE}}\), total time \(T\), and time step \(\Delta t\).
\Ensure Generated video \(\hat{\mathcal{V}}\)

\State \textbf{// Step 1: Initialization}
\State \(t \gets T\)
\State \(\mathbf{z} \gets \mathbf{z}_T\) \Comment{Initialized from $N(0, I)$}

\State \textbf{// Step 2: Euler Integration for Flow Model}
\While{\(t > 0\)}
    \State \(\mathbf{z} \gets \mathbf{z} + f(\mathbf{z}, t)\,\Delta t\) 
    \State \(t \gets t - \Delta t\)
\EndWhile

\State \textbf{// Step 3: Decode the Final Latent to Generate Video}
\State \(\hat{\mathcal{V}} \gets \mathcal{D}_{\text{VAE}}(\mathbf{z})\)

\State \Return \(\hat{\mathcal{V}}\)
\end{algorithmic}
\end{algorithm}

\paragraph{Datasets.}
We validate our experiments using three video datasets: one class-conditional dataset and two unconditional datasets.

\noindent\textbf{UCF-101.}
UCF-101~\citep{soomro2012ucf101dataset101human} is a widely used action recognition dataset that contains 13,320 videos spanning 101 action categories. For our experiments, we use only the training split for model training, ensuring that the test split is not utilized during training. All metric evaluations are conducted on the test split.

\noindent\textbf{SkyTimelapse.} SkyTimelapse is an unconstrained video dataset featuring various time-lapse sequences of the sky, capturing dynamic weather changes and atmospheric conditions. As in ~\citep{yu2023video}, we conduct both training and evaluation on the training split.

\noindent\textbf{FaceForensics.}
FaceForensics~\citep{roessler2019faceforensicspp} is a dataset designed for facial manipulation detection, containing both real and manipulated face videos. We follow the same protocol as SkyTimeLapse, using the training split for both model training and metric evaluations.

\section{V-DiT Architecture}
\label{suppl:architecture}

\begin{figure}[t] 
    \centering
    \includegraphics[width=1.0\columnwidth]{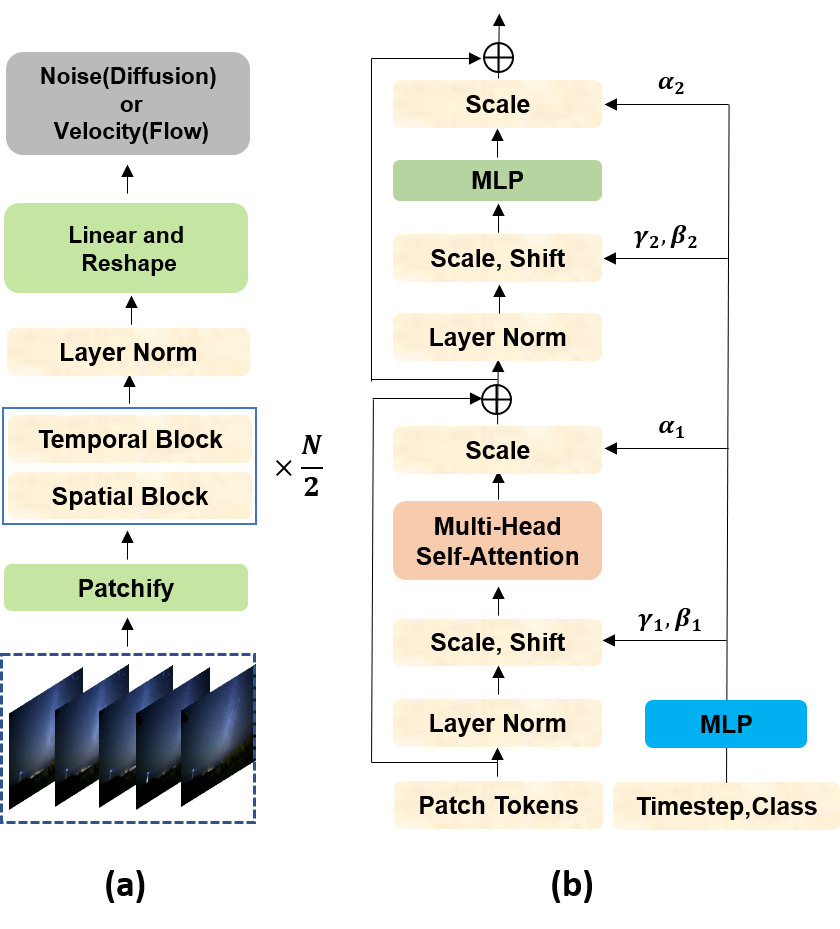} 
    \caption{(a) The V-DiT architecture, consisting of $N$ transformer blocks, each comprising one spatial block and one temporal block. (b) The architecture of a single transformer block.}
    \label{fig:v-dit}
\end{figure}

V-DiT constructs the latent space representation of a video by processing frames individually using the VAE from Stable Diffusion v1.4~\citep{rombach2022high}. Each frame is then transformed into patches via patch embedding, specifically employing a uniform frame patch embedding strategy, where patches are extracted in a frame-wise manner without inter-frame interactions.

The model consists of $N$ transformer blocks, where each unit comprises two sub-blocks: a spatial transformer block and a temporal transformer block. The architecture stacks these units iteratively to build the full transformer model. To encode positional information, spatial positional embeddings and temporal positional embeddings are incorporated before the input is passed into the transformer blocks. Notably, the temporal positional embeddings are applied after the first spatial transformer block and before entering the first temporal transformer block.

Additionally, to incorporate timestep and class information, the model employs adaptive layer normalization (AdaLN), where the scaling parameters $\beta$ and $\gamma$ are learned through a linear projection of the timestep and class embeddings.

\section{Additional Feature Analysis}

\subsection{Additional Assessment of Temporal Stability}
\label{suppl:tracking}
Video encoders such as VideoMAE, which are trained with objectives like masked frame reconstruction and temporal modeling, primarily capture semantic-level dynamics for tasks such as action recognition by aggregating multiple frames into spatio-temporal patches. 
While effective for classification, this design often results in lower discriminability and reduced temporal stability, as also reported in recent zero-shot video segmentation studies~\citep{wang2024zeroshot}. 
Such instability poses challenges for feature alignment in video diffusion models, where a shared MLP projects all tokens regardless of their temporal position. 
In particular, fluctuations in object features across frames can make the alignment optimization process unstable.  

In contrast, image encoders such as DINOv2 and SAM2 better preserve local patterns and maintain consistent spatial representations when applied frame-by-frame, leading to more stable alignment.  
Moreover, PCA-based color consistency serves as a qualitative indicator of temporal coherence; for example, frame-to-frame color shifts can introduce tracking errors (see Fig.~\ref{fig:tracking}).  
Such color-consistency analysis has also been adopted as a practical proxy for evaluating temporal stability in prior works, including TokenFlow~\citep{tokenflow2023} and Chrono PointTrack~\citep{kim2025exploring}.

\subsection{Analysis of Generative Model Feature Representation (SD3)}
\label{suppl:pca-features}

\begin{figure}[t] 
    \centering
    \includegraphics[width=1.0\columnwidth]{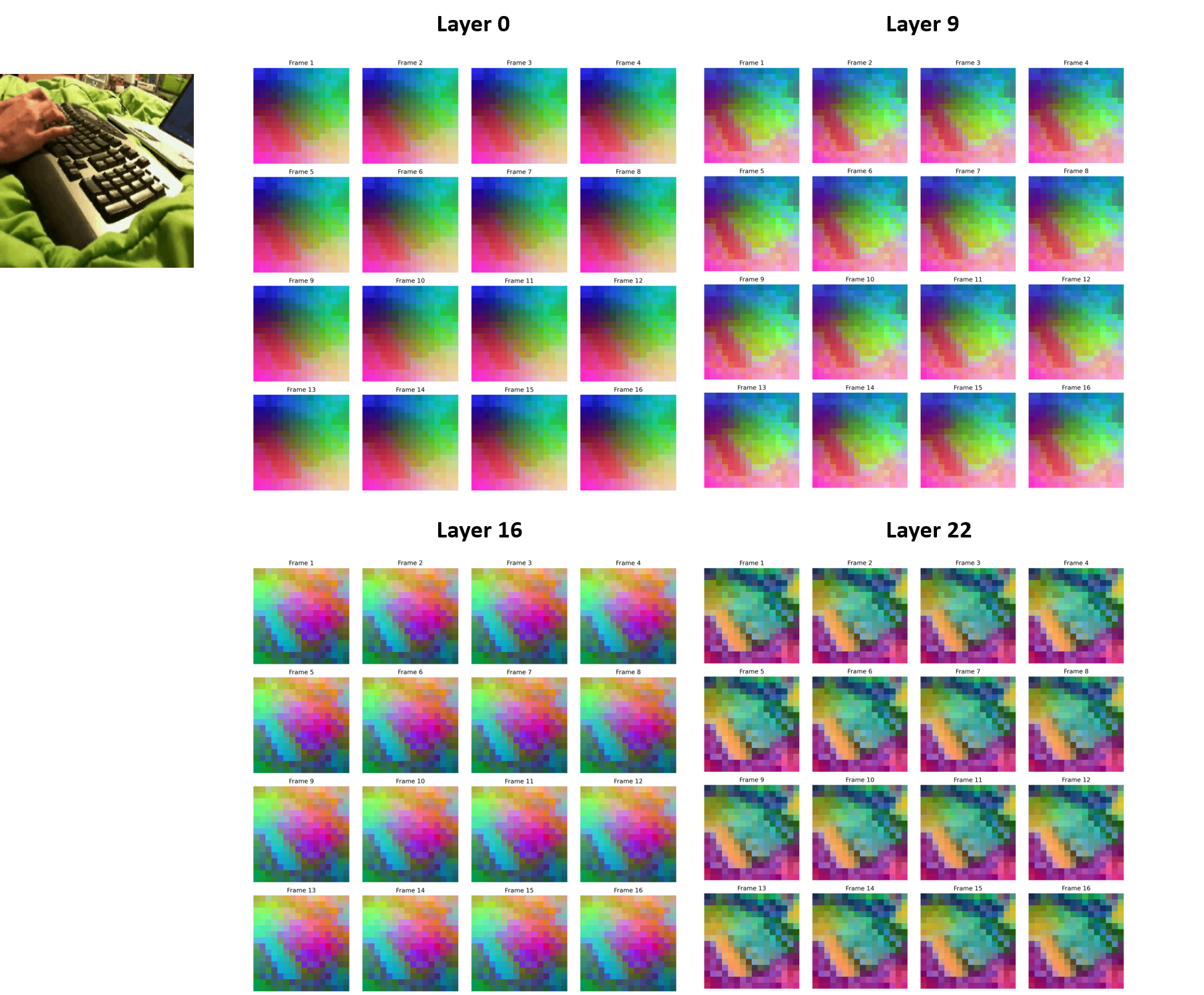} 
    \caption{Based on our PCA analysis of the SD3 transformer block, we observed that the output following the 22th transformer block is the most discriminative and retains high-frequency details. Therefore, we have decided to use this feature for alignment.}
    \label{fig:pca_sd3}
\end{figure}

We conducted a PCA analysis of features extracted from the transformer blocks of SD3 at different layer depths. The results showed that as the layer depth increased, the extracted information became more detailed and well-defined. Based on the PCA results in Fig.~\ref{fig:pca_sd3}, we identified that the output features from the 22th transformer block exhibited the highest degree of object distinction and consistency. Therefore, we selected the features from this specific layer in Tab~\ref{tab:ablation_study different encoders}.

\begin{figure}[t]
    \centering
    \newcommand{\columnWidthFraction}{0.24\columnwidth}
    \newcommand{\videoHeight}{2.0cm}
    \begin{tabular}{@{}c@{}c@{}c@{}c@{}}
        \begin{minipage}{\columnWidthFraction}
            \centering
            \animategraphics[loop, width=\linewidth]{8}{rebuttal_videos/jumprope/}{00}{47}
        \end{minipage} &
        \begin{minipage}{\columnWidthFraction}
            \centering
            \animategraphics[loop, width=\linewidth]{8}{rebuttal_videos/jumprope_pca/}{00}{47}
        \end{minipage} &
        \begin{minipage}{\columnWidthFraction}
            \centering
            \includegraphics[width=\linewidth, height=\videoHeight]{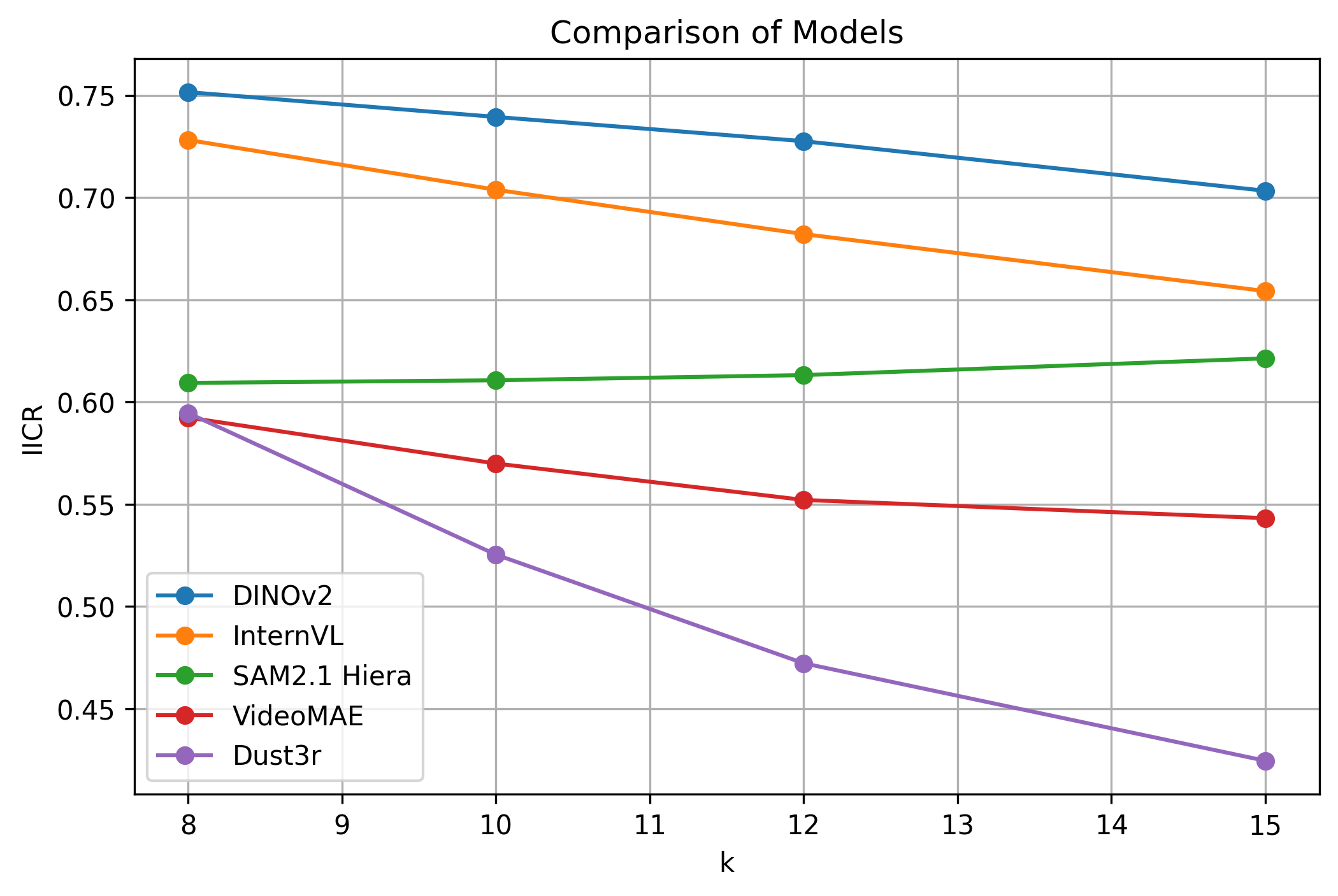}
        \end{minipage} &
        \begin{minipage}{\columnWidthFraction}
            \centering
            \tiny 
            \renewcommand{\arraystretch}{1.1}
            \begin{tabular}{@{}l r@{}}
                \toprule
                Method & FVD \\
                \midrule
                V-DiT & 412.21 \\
                + DINOv2 & \textbf{347.29} \\
                + SAM2.1 & 357.70 \\
                + Fusion & 347.45 \\
                + InternVL & 360.93 \\
                \bottomrule
            \end{tabular}
        \end{minipage}
    \end{tabular}

    \caption{ \footnotesize
        From left to right: input video, PCA visualization using InternVL features, IICR metric plot, and FVD scores.
    }
    \label{fig:mllm}
\end{figure}

\subsection{Analysis of MLLM Feature Representation}
Recently, multi-modal large language models (MLLMs) have emerged as a rapidly growing research direction, with vision encoders playing a crucial role in their performance. 
Motivated by this trend, we also analyzed the vision encoder used in a state-of-the-art MLLM to better understand its potential for video generation tasks. 
In particular, we evaluated InternVL, a leading MLLM, for discriminative power and temporal consistency using the proposed IICR metric and PCA analysis, as shown in Fig.~\ref{fig:mllm}-left. 
Although InternVL features are less discriminative than those of DINOv2, they still exhibit reasonable performance. 
However, they introduce a substantial computational overhead, running approximately $5\times$ slower than DINOv2. 
Due to time constraints, we trained InternVL-based models for up to 100K iterations and compared them with baselines, as shown in Fig.~\ref{fig:mllm}-right. 
While achieving competitive performance, this approach faces practical limitations in terms of training efficiency.

\section{Limitation}
\label{suppl:limitations}

\paragraph{Extending Text-to-Image to Text-to-Video.}

For text-to-video generation, we implement a larger-scale V-DiT model using OpenDiT, resulting in 28 transformer blocks with 760M parameters. The spatial layers are initialized with text-to-image pre-trained weights from PixArt-$\alpha$~\citep{chen2023pixartalpha}. Training is conducted on the OpenVid-1M dataset~\citep{nan2024openvid}, where videos have a resolution of $256 \times 256 \times 16$. To capture longer-range motion, we sample video data with a frame stride of 3 and train the model for 70K steps with a total batch size of 128. Also, we set the patch size to $2 \times 2$ and fix the alignment loss coefficient $\gamma$ at 0.5. 

Despite these efforts, we encountered several challenges. In particular, the feature projection loss (projection loss) initially struggled to align the image diffusion model with the video model, leading to poor projection in the early training stages. However, as training progressed, the projection score improved beyond 0.6, indicating better alignment. Nevertheless, this improvement came at a cost: the spatial transformer blocks deviated from their original logic, which ultimately led to a decline in overall performance. We observed that while feature projection can accelerate semantic learning when training from scratch, it interferes with models that already possess strong semantic priors, potentially disrupting pre-learned representations. 

Given computational constraints, we were unable to train the text-to-video model entirely from scratch, which would have provided a cleaner evaluation of semantic learning dynamics. Therefore, as part of our future work, we will try applying our method to a text-to-video model from scratch.

\begin{table}[t]
\centering
\renewcommand{\arraystretch}{1.2}
\resizebox{\columnwidth}{!}{%
\begin{tabular}{lcccc}
\toprule
Method & Latte (V-DiT) & V-DiT + Fusion & PVDM & Matten \\
\midrule
FVD $\downarrow$ & 180.06 & \textbf{154.21} & 493.67 & 210.61 \\
\bottomrule
\end{tabular}
}
\caption{FVD comparison of different methods. 
We report standard I3D-based FVD without content-debiased evaluation.}
\label{tab:comp_baseline}
\end{table}

\section{Comparison to Previous Methods}
To rigorously validate the effectiveness of our method, it is ideal to conduct training under identical conditions (e.g., batch size, learning rate, etc.). However, prior works adopt varying batch sizes and learning rates, making a direct comparison inherently challenging. For fairness, we nonetheless report Fréchet Video Distance (FVD) values for all methods. Specifically, for Latte, we re-trained the model using our experimental settings; for PVDM, we used the publicly released checkpoint; and for Matten, which is not publicly available, we report the values provided in its original paper. Moreover, since none of the compared methods employed content-debiased FVD, we evaluate all methods using the standard I3D-based FVD metric. Our approach not only enhances the original baseline but also achieves superior performance compared to all other baselines as Tab~\ref{tab:comp_baseline}.

\section{Additional Qualitative Results}
\label{suppl:qualitative}

We provide additional qualitative comparison results in Fig. \ref{fig: supp qual1}, \ref{fig: supp qual2}, and \ref{fig: supp qual3}. Please visit our project page for full video results.

\begin{figure*}[t]
    \centering
    \newcommand{\numColumns}{1}
    \newcommand{\columnSpacing}{0.1cm}

    \begin{tabularx}{\textwidth}{X}
        \centering \small \textbf{V-DiT}
    \end{tabularx}

    \begin{tabular}{
        @{}
        p{\dimexpr(\textwidth-\columnSpacing*(\numColumns-1))/\numColumns} @{}
    }
        \animategraphics[loop, width=\linewidth]{8}{videos/video_origin/}{00}{15}
    \end{tabular}

    \vspace{1.0em}

    \begin{tabularx}{\textwidth}{X}
        \centering \small \textbf{V-DiT + Ours(\emph{Fusion})}
    \end{tabularx}

    \begin{tabular}{
        @{}
        p{\dimexpr(\textwidth-\columnSpacing*(\numColumns-1))/\numColumns} @{}
    }
        \animategraphics[loop, width=\linewidth]{8}{videos/video_ours/}{00}{15}
    \end{tabular}

    \vspace*{0.2cm}
    \caption{
    More qualitative results on the UCF-101 dataset for V-DiT versus V-DiT + Ours trained on 1M iterations, clearly demonstrating enhanced image quality and superior motion dynamics. \textit{Best viewed with Acrobat Reader. Click image to play the video clip.}
    }
    \label{fig: supp qual1}
\end{figure*}

\begin{figure*}[t]
    \centering
    \newcommand{\numColumns}{1}
    \newcommand{\columnSpacing}{0.1cm}

        \centering
        \begin{tabularx}{\textwidth}{X}
            \centering \small \textbf{V-DiT}
        \end{tabularx}

        \begin{tabular}{
            @{}
            p{\dimexpr(\textwidth-\columnSpacing*(\numColumns-1))/\numColumns} @{}
        }
            \animategraphics[loop, width=\linewidth]{8}{videos/video_ffs_origin/}{00}{15}
        \end{tabular}

        \vspace{1.0em}

        \begin{tabularx}{\textwidth}{X}
            \centering \small \textbf{V-DiT + Ours(\emph{Fusion})}
        \end{tabularx}

        \begin{tabular}{
            @{}
            p{\dimexpr(\textwidth-\columnSpacing*(\numColumns-1))/\numColumns} @{}
        }
            \animategraphics[loop, width=\linewidth]{8}{videos/video_ffs_ours/}{00}{15}
        \end{tabular}

        \caption{Qualitative results on the FaceForensics dataset for V-DiT versus V-DiT + Ours trained on 200K iterations, demonstrating enhanced image quality with better-defined structure.}

    \hfill
        \centering
        \begin{tabularx}{\textwidth}{X}
            \centering \small \textbf{V-DiT}
        \end{tabularx}

        \begin{tabular}{
            @{}
            p{\dimexpr(\textwidth-\columnSpacing*(\numColumns-1))/\numColumns} @{}
        }
            \animategraphics[loop, width=\linewidth]{8}{videos/video_sky_origin/}{00}{15}
        \end{tabular}

        \vspace{1.0em}

        \begin{tabularx}{\textwidth}{X}
            \centering \small \textbf{V-DiT + Ours(\emph{Fusion})}
        \end{tabularx}

        \begin{tabular}{
            @{}
            p{\dimexpr(\textwidth-\columnSpacing*(\numColumns-1))/\numColumns} @{}
        }
            \animategraphics[loop, width=\linewidth]{8}{videos/video_sky_ours/}{00}{15}
        \end{tabular}

        \caption{Qualitative results on the SkyTimeLapse dataset for V-DiT versus V-DiT + Ours trained on 200K iterations.}

    \vspace*{0.2cm}
    \caption{
    More qualitative results comparing V-DiT and V-DiT + Ours (\emph{Fusion}) on unconditional video datasets. \textit{Best viewed with Acrobat Reader. Click image to play the video clip.}
    }
    \label{fig: supp qual2}
\end{figure*}

\begin{figure}[t]
    \centering
    \newcommand{\numColumns}{1}
    \newcommand{\columnSpacing}{0.1cm}

    \begin{tabularx}{\columnwidth}{X}
        \centering \small \textbf{First Row of Fig.~\ref{fig:main_fig}}
    \end{tabularx}

    \begin{tabular}{
        @{}
        p{\dimexpr(\textwidth-\columnSpacing*(\numColumns-1))/\numColumns} @{}
    }
        \animategraphics[loop, width=\columnwidth]{8}{videos/row_1/}{00}{15}
    \end{tabular}

    \vspace{1.0em}

    \begin{tabularx}{\columnwidth}{X}
        \centering \small \textbf{Second Row of Fig.~\ref{fig:main_fig}}
    \end{tabularx}

    \begin{tabular}{
        @{}
        p{\dimexpr(\textwidth-\columnSpacing*(\numColumns-1))/\numColumns} @{}
    }
        \animategraphics[loop, width=\columnwidth]{8}{videos/row_2/}{00}{15}
    \end{tabular}

    \vspace{1.0em}

    \begin{tabularx}{\columnwidth}{X}
        \centering \small \textbf{Third Row of Fig.~\ref{fig:main_fig}}
    \end{tabularx}

    \begin{tabular}{
        @{}
        p{\dimexpr(\textwidth-\columnSpacing*(\numColumns-1))/\numColumns} @{}
    }
        \animategraphics[loop, width=\columnwidth]{8}{videos/row_3/}{00}{15}
    \end{tabular}

    \vspace{1.0em}

    \begin{tabularx}{\columnwidth}{X}
        \centering \small \textbf{Fourth Row of Fig.~\ref{fig:main_fig}}
    \end{tabularx}

    \begin{tabular}{
        @{}
        p{\dimexpr(\textwidth-\columnSpacing*(\numColumns-1))/\numColumns} @{}
    }
        \animategraphics[loop, width=\columnwidth]{8}{videos/row_4/}{00}{15}
    \end{tabular}

    \vspace{1.0em}

    \begin{tabularx}{\columnwidth}{X}
        \centering \small \textbf{Fig.~\ref{fig:main_ffs}}
    \end{tabularx}

    \begin{tabular}{
        @{}
        p{\dimexpr(\textwidth-\columnSpacing*(\numColumns-1))/\numColumns} @{}
    }
        \animategraphics[loop, width=\columnwidth]{8}{videos/row_5/}{00}{15}
    \end{tabular}

    \vspace*{0.2cm}
    \caption{
    Qualitative video result of Fig.~\ref{fig:main_fig} and Fig.~\ref{fig:main_ffs}. \textit{Best viewed with Acrobat Reader. Click image to play the video clip.}
    }
    \label{fig: supp qual3}
\end{figure}

\end{document}